\documentclass[]{tokenwave}

\usepackage[utf8]{inputenc}
\usepackage{url}
\usepackage{xurl}
\usepackage{amsfonts,amsmath,amssymb}
\usepackage{nicefrac}
\usepackage{array}
\usepackage{tikz}
\usetikzlibrary{positioning, arrows.meta, fit, backgrounds, calc}

\crefname{figure}{Figure}{Figures}
\Crefname{figure}{Figure}{Figures}
\crefname{table}{Table}{Tables}
\Crefname{table}{Table}{Tables}
\crefname{section}{Section}{Sections}
\Crefname{section}{Section}{Sections}
\crefname{subsection}{Section}{Sections}
\Crefname{subsection}{Section}{Sections}
\crefname{appendix}{Appendix}{Appendices}
\Crefname{appendix}{Appendix}{Appendices}
\crefname{equation}{Equation}{Equations}
\Crefname{equation}{Equation}{Equations}

\titleformat*{\paragraph}{\sffamily\bfseries}

\newcommand{\cmark}{$\checkmark$}
\newcommand{\pmark}{$\sim$}
\newcommand{\xmark}{$\times$}

\title{REE:  A Scientific Research Entrance Exam for AI Agents}

\author[1,2]{Zeqi Zhou}
\author[2]{Minghao Liu}
\author[3]{Jiarui Liu}
\author[4]{Qingcheng Zeng}
\author[5]{Rui Yang}
\author[6]{Tianyu Liu}
\author[1,7]{Zetao Yang}
\author[1]{Wenqiang Zu}
\author[8]{Enduo Zhao}
\author[9]{Heng-Jui Chang}
\author[10]{Lun-Hsin Kuo}
\author[11]{Meng-Fu Chen}
\author[12]{Haokai Zhao}
\author[6]{Weijia Zhang}
\author[13]{Jen-Tse Huang}
\author[1]{Stephen Huang}
\author[1]{Eli Cheung}
\author[14]{Zhijing Jin}
\author[3]{Mona Diab}
\author[1,2]{Weihao Xuan}

\affiliation[1]{TokenWave AI\protect\textsuperscript{\protect\textdagger}}
\affiliation[2]{The University of Tokyo}
\affiliation[3]{Carnegie Mellon University}
\affiliation[4]{Northwestern University}
\affiliation[5]{Weill Cornell Medicine}
\affiliation[6]{Yale University}
\affiliation[7]{Nanyang Technological University}
\affiliation[8]{Monolith Management, US}
\affiliation[9]{Stanford University}
\affiliation[10]{California Institute of Technology}
\affiliation[11]{Harvard University}
\affiliation[12]{University of New South Wales}
\affiliation[13]{Johns Hopkins University}
\affiliation[14]{University of Toronto}

\titlelogoplacement{top}
\renewcommand{\titlelogoheight}{0.74cm}
\titlelogo{figures/logos/tokyo}
\titlelogo{figures/logos/cmu}
\titlelogo{figures/logos/northwestern}
\titlelogo{figures/logos/weill-cornell}
\titlelogo{figures/logos/yale}
\titlelogo{figures/logos/ntu}
\titlelogo{figures/logos/stanford}
\titlelogo{figures/logos/caltech}
\titlelogo{figures/logos/harvard}
\titlelogo{figures/logos/unsw}
\titlelogo{figures/logos/jhu}
\titlelogo{figures/logos/toronto}
\titlelogo{figures/logos/monolith}

\abstract{Autonomous research agents are emerging as a promising approach to scientific problem-solving across disciplines, owing to their growing ability to improve a solution through repeated experiments. Evaluating this ability requires benchmarks that pose open-ended research problems and measure improvements over established scientific methods. We introduce REE (Research Entrance Exam), a benchmark that tests AI agents on scientific research. It consists of 100 open-ended research problems spanning five broad scientific areas and 23 domains. To ensure high quality, every problem is drawn from active research in its field and individually reviewed by domain experts. Each problem is posed as a task of iterative improvement: an agent has up to 12 hours to improve a working starter solution, and REE scores the result on a common scale against an established expert solution evaluated under comparable conditions. By this measure, two of the four frontier agents we evaluate, Claude Opus 5.5 and GPT-6.1 Sol, outperform the expert solutions on 86\% and 81\% of problems, across all five scientific areas. Their runs also show how they do so: both agents surpass the expert solutions within the first two hours on average, and their best solutions mostly tailor or combine established methods of the field. REE thus provides a unified framework for evaluating autonomous scientific research, capturing both improvements over established methods and the research processes behind them.
}

\twinfo[Data]{\url{https://huggingface.co/datasets/Tokenwave-AI/REE}}

\begin{document}

\maketitle
\begingroup\renewcommand{\thefootnote}{\textdagger}\footnotetext{Located in Palo Alto, United States.}\endgroup

\section{Introduction}
\label{sec:intro}

Autonomous research agents are AI systems that propose methods, run experiments, and refine their solutions to a research problem with little human guidance. Their role in scientific research is growing, from supporting single steps, such as searching the literature or writing code, to carrying out this whole cycle. Results produced in this way have begun to hold up under experimental validation and peer review \citep{gottweis2025coscientist,yamada2025aiscientistv2}. The longer-term aim is for such agents to work across scientific disciplines and advance the frontier problems of each.

One of the most effective modes of agent research is iterative improvement. The agent starts from a working solution and a measurable objective, proposes a change, tests it, keeps the change if the result improves, and repeats this cycle many times. AlphaEvolve applied this cycle to more than fifty open problems in analysis, geometry, combinatorics, and number theory, and it improved the best known solution on 20\% of them \citep{deepmind2025alphaevolveblog,novikov2025alphaevolve}. The cycle is also effective in direct competition with humans: in a four-hour AtCoder Heuristic Contest, where participants refine a program against a score, an agent took first place ahead of 804 human participants \citep{sakana2025ahc058}. In both cases the object being improved is a solution to a given problem. When it is the agent itself, the cycle is known as recursive self-improvement (RSI), a direction that now draws wide attention because an agent that improves itself could speed up all of its later work \citep{srikanth2026aide2}.

Evaluation is moving toward iterative improvement as well, most clearly in machine learning and computing, where a growing number of benchmarks measure open-ended improvement over hours in place of a single attempt, some of them explicitly as a step toward RSI \citep{wijk2025rebench,chan2025mlebench,imajuku2025alebench,meng2026rsibenchdata,chi2026ai4aibench}. For scientific problems, however, evaluation has not yet followed to the same extent. Most benchmarks in the sciences still check an agent's output against a known answer or a pass criterion \citep{tian2024scicode,chen2025scienceagentbench}, and the few that measure improvement cover a limited range of scientific domains \citep{wang2026naturebench,agent4sci2026}. Measuring iterative improvement across the sciences has therefore received little attention so far.

\begin{figure}[t]
  \centering
  \includegraphics[width=\linewidth]{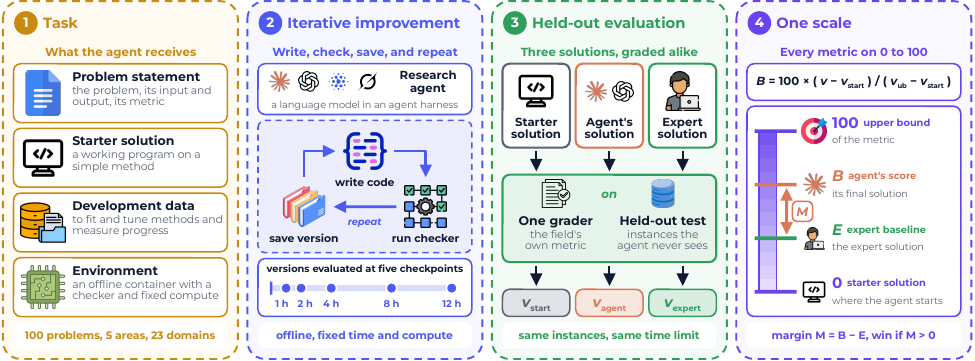}
  \caption{An REE task in four steps. (1) The agent receives a problem statement, a starter solution, development data, and an offline environment with a checker. (2) It improves the solution iteratively for up to 12 hours, and its saved versions are evaluated at five checkpoints. (3) One grader evaluates the starter solution, the agent's solution, and the expert solution on the same held-out test instances. (4) Each metric value is mapped to a score from 0, the starter solution, to 100, the upper bound of the metric. The margin $M = B - E$ compares the agent's score $B$ with the expert baseline $E$.}
  \label{fig:task}
\end{figure}

We introduce REE (Research Entrance Exam), a benchmark for AI agents that measures iterative improvement on 100 open-ended research problems across the sciences (\Cref{fig:task}). The problems cover 23 domains in five scientific areas: mathematics and computing, physical sciences and engineering, life and health sciences, earth and environmental sciences, and language, society and culture. We call REE an entrance exam because it asks an agent to do what a field expects of a new researcher, which is to take one of the field's problems and improve on its established solution. Accordingly, each problem is one that its scientific community poses, evaluates, and continues to study, it keeps the objective and the evaluation metric of its field, and it has been reviewed by domain experts. The problems are open-ended in that a solution is a program scored by how far it improves a quantitative metric, and it is never simply marked right or wrong.

To evaluate an agent, REE poses each of these problems as a task of iterative improvement. The agent receives a problem statement, a working starter solution, and development data on which it can develop and test its methods. It then has up to 12 hours to improve the solution, and a provided checker scores each revision on the development data. The final solution is evaluated on held-out test instances that the agent never sees. It receives a score from 0 to 100, where 0 is the starter solution and 100 is the upper bound of the problem's metric, such as the highest F-score that any output can attain against the labels of two annotators. To relate this score to human work, each problem also has an expert solution, an established human-designed solution such as a competition entry or a method published for the problem. Because published results were obtained on other test data, we evaluate each expert solution ourselves on the same test instances and under comparable conditions. We also evaluate the intermediate solutions that an agent saves during a run, which shows how its progress develops over time.

Under this protocol, we evaluate four frontier agents: Claude Opus 5.5, GPT-6.1 Sol, GLM-5.3, and Grok 4.7, each a language model running in an agent harness. The expert solutions score 24.4 on average, and the two leading agents, Claude Opus 5.5 and GPT-6.1 Sol, score 39.6 and 38.2. They outperform the expert solution on 86\% and 81\% of problems, and their mean score is above that of the expert solutions in each of the five scientific areas. The intermediate solutions and the final methods show how these results come about. On average, both leading agents surpass the expert solutions within two hours and reach 87\% of their final score within four hours. Their best solution on each problem mostly builds on existing methods, tailoring the standard method of the field on 51\% of problems and combining existing methods on 45\%. The other two agents, GLM-5.3 and Grok 4.7, score 28.6 and 25.3, close to the expert solutions, which shows that REE separates current agents. It also leaves room to measure stronger ones, as even the highest mean score, 39.6, is less than halfway to the upper bound of 100. In summary, our contributions are as follows.
{\setlength{\leftmargini}{1.4em}\begin{itemize}\setlength{\itemsep}{1pt}\setlength{\parskip}{0pt}
  \item \textbf{A Benchmark of Open-Ended Research Problems across the Sciences.} REE extends the evaluation of iterative improvement to the sciences, with 100 open-ended research problems in five scientific areas and 23 domains, each reviewed by domain experts. We release all problems with their data, graders, and starter and expert solutions.
  \item \textbf{A Common Scale with a Human Reference.} REE scores every agent on one scale shared by all problems and against an expert solution evaluated under comparable conditions, and it follows each run over time. Results are therefore comparable across fields, with human work, and over the course of a run.
  \item \textbf{Findings on Frontier Research Agents.} The leading agents already outperform the expert solutions on most problems and in every scientific area. They do so early in a run and mostly by building on existing methods, and they remain far from the upper bound.
\end{itemize}}

\section{Related Work}
\label{sec:related}

\begin{table}[t]
\caption{REE combines a human reference evaluated under comparable conditions, held-out evaluation, and evaluation over time, and it covers 23 scientific domains. \emph{Horizon}: wall-clock time of one agent run. \emph{Human reference}: the human result that agents are compared with. \cmark{} means it is evaluated under conditions comparable to the agent's, and Agent4Sci's \pmark{} marks a published best model that is re-run as the agent's starting point. \emph{Held-out}: the test instances are hidden from the agent, and it receives no score on them during the run. \emph{Over time}: solutions are also evaluated at intermediate times. \cmark{} yes, \pmark{} in part, \xmark{} no, -- not applicable.}
\label{tab:related}
\centering
\scriptsize
\setlength{\tabcolsep}{3pt}
\begin{tabular}{@{}lr>{\raggedright\arraybackslash}p{0.14\linewidth}l>{\raggedright\arraybackslash}p{0.25\linewidth}cc@{}}
\toprule
Benchmark & Tasks & Fields & Horizon & Human reference & Held-out & Over time \\
\midrule
MLE-bench \citep{chan2025mlebench} & 75 & ML engineering on Kaggle problems from many fields & 24\,h & \pmark{} Kaggle leaderboard, recorded & \pmark & \pmark \\
RE-Bench \citep{wijk2025rebench} & 7 & ML R\&D & 8\,h & \cmark{} live experts, same VMs & \pmark & \cmark \\
PaperBench \citep{starace2025paperbench} & 20 & ML & 12\,h & \pmark{} 8 ML PhDs on 4 papers & -- & \pmark \\
MLRC-Bench \citep{zhang2025mlrcbench} & 7 & ML & 5--10\,h & \pmark{} top entry, recorded or re-run & \cmark & \pmark \\
ALE-Bench \citep{imajuku2025alebench} & 40 & Heuristic optimization & 4\,h & \pmark{} contestants, recorded & \cmark & \pmark \\
AIRS-Bench \citep{lupidi2026airsbench} & 20 & ML & 24\,h & \pmark{} literature state of the art, quoted & \pmark & \pmark \\
NatureBench \citep{wang2026naturebench} & 90 & 6 science domains & 4\,h & \pmark{} paper state of the art, quoted & \pmark & \pmark \\
Agent4Sci \citep{agent4sci2026} & 25 & 11 science domains & 12\,h & \pmark{} published best, re-run as the start & \cmark & \pmark \\
FrontierCS~2 \citep{mang2026frontiercs2} & 48 & 13 CS domains & 5\,h & \cmark{} authors' code, same held-out suite & \cmark & \pmark \\
LHTB~2.0 \citep{li2026lhtb2} & 77 & 13 domains & 3\,h & \xmark & \cmark & \cmark \\
RSI-Exam \citep{rsiexam2026} & 88 & 6 domains & 12\,h & \pmark{} reference set to 0.6, where one exists & \cmark & \pmark \\
\midrule
\textbf{REE (ours)} & \textbf{100} & \textbf{23 domains} & \textbf{12\,h} & \textbf{\cmark{} expert solution, comparable conditions} & \textbf{\cmark} & \textbf{\cmark} \\
\bottomrule
\end{tabular}
\end{table}

\paragraph{Benchmarks for research agents.} Long-horizon benchmarks for research agents have so far centered on machine learning. RE-Bench evaluates agents on seven open-ended machine-learning research engineering environments \citep{wijk2025rebench}, and MLE-bench draws its tasks from Kaggle competitions \citep{chan2025mlebench}. Related benchmarks extend this setting to heuristic optimization and computer science \citep{imajuku2025alebench,mang2026frontiercs2}, and a recent group frames the improvement of training data, training algorithms, or working methods as a step toward recursive self-improvement \citep{meng2026rsibenchdata,chi2026ai4aibench,rsiexam2026}. Benchmarks on scientific problems, in contrast, have mostly checked an agent's output against a known answer or a pass criterion \citep{tian2024scicode,chen2025scienceagentbench,majumder2024discoverybench,phan2026hle,glazer2024frontiermath}. Only recently have long-horizon benchmarks begun to move toward the sciences. Agent4Sci covers 25 tasks in 11 scientific domains \citep{agent4sci2026}, and NatureBench builds 90 tasks in six domains from Nature-family papers \citep{wang2026naturebench}. REE extends this breadth further, to 100 problems in five areas and 23 domains, including language, society and culture.

\paragraph{Human reference.} The human results that benchmarks compare agents with are of three kinds (\Cref{tab:related}). The first is a recorded competition result, such as the Kaggle leaderboards in MLE-bench and the contest rankings in ALE-Bench \citep{chan2025mlebench,imajuku2025alebench}. The second is a result quoted from the literature, such as the published state of the art in AIRS-Bench and NatureBench \citep{lupidi2026airsbench,wang2026naturebench}. In both, the human number was obtained with data, time, and compute that differ from the agent's. The third kind is a human result produced under the agent's conditions, which removes this difference. RE-Bench runs human experts on the same machines \citep{wijk2025rebench}, and FrontierCS~2 evaluates the code released with a paper on the same held-out suite \citep{mang2026frontiercs2}. Agent4Sci also re-runs the published best model, but it uses this model as the agent's starting point rather than as a separate reference \citep{agent4sci2026}. REE belongs to the third kind and provides such a reference for every problem, as an established human-designed solution that is evaluated separately from the agent under comparable conditions. \Cref{app:related} compares the benchmarks in more detail, including their scoring scales and their evaluation over time.

\section{The REE Benchmark}
\label{sec:benchmark}

REE measures how far a research agent can improve a solution to a scientific problem, and how the result compares with the established work of human experts. Every task follows the same design. The agent is given an open-ended research problem and a working solution to start from, and it has a fixed budget to improve that solution by any method it chooses. REE then reports the result on a scale shared by all problems, next to an expert solution evaluated under comparable conditions, which gives every score a human reference. The rest of this section describes the components of this design: the problems, the task format, the metrics, and the composition of the benchmark.

\subsection{Problems}
\label{sec:problems}

REE's problems are research problems that scientific communities pose, evaluate, and continue to study. From such problems, we selected those that meet three criteria. The first is that the problem is open-ended, which means that a solution is a program scored by a quantitative metric over many instances, and it is never simply marked right or wrong. The second is that it can be evaluated automatically on instances that the agent has not seen, and the third is that it has an established human-designed solution that can be run within the budget of the task. Each selected problem keeps the objective and the evaluation metric that its field uses, so a solution is judged by the field's own measure. In building gene trees from protein sequences, for example, a solution is scored by the number of branchings that differ from a reference tree, which a generic distance loss does not capture. Finally, domain experts took part in posing the problems and reviewed every one of them, checking that the problem is well posed and that it reflects the current state of research in its field. \Cref{app:hard} illustrates on five tasks what makes the problems challenging.

\subsection{Task Format and Evaluation}
\label{sec:day}

Each task gives the agent a problem statement, a starter solution, and development data, and it holds out a separate set of test instances for evaluation. The statement defines the problem, its input and output, and its metric. The starter solution is a complete, working program built on a short, simple method, and the agent may improve, replace, or discard it. The development data serve to fit and tune methods and to measure progress. In one task, for example, the statement asks for the verbal multiword expressions in English sentences and defines an F1 score as the metric, the starter solution reaches an F1 of 0.319 on the held-out test instances, and the development data are sentences with annotated expressions.

A run lasts up to 12 hours within a fixed compute budget. The agent works offline, so that it cannot retrieve a published solution or answer during the run (\Cref{app:runs}). In this time it writes and runs code, saves versions of its solution, and decides when to stop. A checker is available throughout: it executes the current solution on the development data under the evaluation conditions and reports the score, so the agent can confirm that the solution runs and can measure the effect of each change.

After the run, the final solution is evaluated in a separate container on the held-out test instances, which the agent never sees and on which it receives no score during the run. On some tasks these instances also come from subjects or sites that the development data do not contain, such as recordings from new patients in seizure detection. A score on the held-out test instances therefore measures how well a method generalizes beyond the development data. One grader scores the starter solution, the agent's solution, and the expert solution on the same instances. The evaluation uses the time limit given in the task's statement, and an instance on which a solution fails, exceeds this limit, or writes invalid output receives the failure value defined there.

REE also evaluates each run over time, using the time at which each version of the solution was saved. After the run, the same grader evaluates the version that was current at 1, 2, 4, and 8 hours on the held-out test instances, and the agent sees none of these scores. Together with the final solution, these four versions give the score of a run at five checkpoints.

\subsection{Metrics}
\label{sec:scale}

REE reports two quantities for a solution. The \emph{score} measures how far the solution moves a task's metric from the starter solution toward the upper bound of the metric. The \emph{margin} compares this score with the score of an expert solution on the same task. Both quantities rest on the upper bound, which we define first.

Each task has an upper bound $v_{\mathrm{ub}}$ on its metric, which is set in one of three ways. For 89 tasks it is the best value that the metric can take by definition, such as an F1 of 1 or an error of zero, whether or not a method can reach this value on the task's data. For 7 tasks a bound can be computed for the problem itself, and REE uses it in place of the definitional value. In sound event detection, for example, two annotators labeled each recording independently, and the upper bound is the highest mean F-score that any output can reach against both sets of annotations, which is 0.957. For the remaining 4 tasks the metric has no finite or usable best value, as with a ratio in decibels, and the upper bound is an estimate of the best attainable value, based on the performance of current state-of-the-art methods on the problem (\Cref{app:measures} notes two special cases).

Each field scores its problems in its own units, such as an error in millimoles per gram, an F1 score, a count of comparators, or a ratio in decibels. To make results comparable across tasks, REE maps the metric of every task to a score from 0 to 100, on which 0 is the starter solution and 100 is the upper bound. For a solution with metric value $v$ on the held-out test instances, the score is
\begin{equation}
  B \;=\; 100 \cdot \frac{v - v_{\mathrm{start}}}{v_{\mathrm{ub}} - v_{\mathrm{start}}},
  \label{eq:scale}
\end{equation}
where $v_{\mathrm{start}}$ is the value of the starter solution on the same instances, and scores are clipped to $[0, 100]$. In the multiword-expression task above, the metric is the F1 score, so the upper bound is 1, and the starter solution reaches 0.319. A solution with an F1 of 0.512 therefore receives a score of $100 \cdot (0.512 - 0.319) / (1 - 0.319) \approx 28$.

To relate the score to human work, each task also has an expert solution, an established human-designed solution to the problem. It is, for example, an entry in the problem's organized evaluation or a method published for the problem, and \Cref{app:anchor-rule} lists the five kinds with the number of tasks of each. The expert solution is independent of the starter solution, and the agent does not see it. Its published result cannot be placed on the REE scale directly, because it was obtained on other test data and often with more data, time, or compute than a task provides. REE therefore runs each expert solution itself, offline, on the same held-out test instances, within the same 12 hours, and with at least the compute that the agent's code runs on. The resulting score is the \emph{expert baseline} $E$, which can differ from the performance reported in the method's original publication.

For an agent with score $B$ on a task, the margin is $M = B - E$, and a positive margin means that the agent's final solution scores higher than the expert solution. Over all tasks, we report the mean score, the mean margin, and the \emph{win rate}, which is the share of tasks with a positive margin.

\subsection{Composition}
\label{sec:composition}

\begin{figure}[t]
\centering
\includegraphics[width=\linewidth]{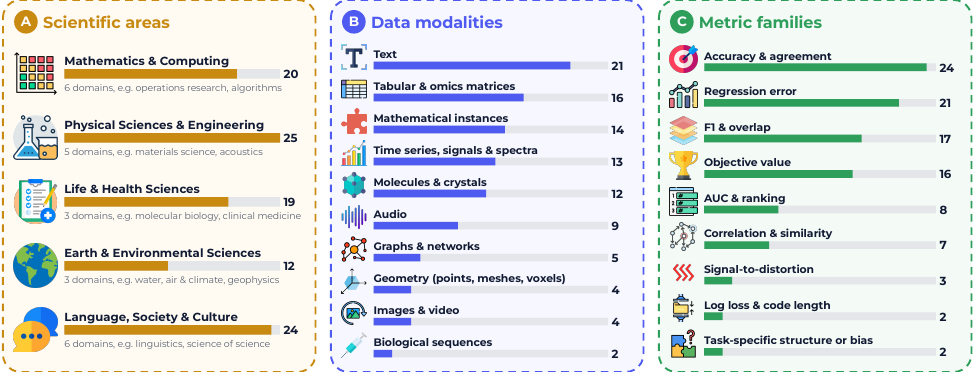}
\caption{REE covers 100 tasks in five areas and 23 domains, with 10 data modalities and 9 metric families. (a) Tasks per scientific area, with the number of domains in each area. (b) Tasks per data modality. (c) Tasks per metric family.}
\label{fig:composition}
\end{figure}

REE's 100 tasks are grouped into five areas, 23 domains, and 91 sub-fields (\Cref{fig:composition,tab:composition}). The domains include materials science, acoustics, molecular biology, clinical medicine, linguistics, and operations research. The tasks cover a range of problem types, including classification, regression, structured prediction, inverse problems, and combinatorial optimization. They use 10 data modalities, such as text, time series, molecules, and audio, and their metrics fall into 9 families. The same breadth holds within a single domain: acoustics alone includes classification, structured prediction, and inverse problems.

{\newcommand{\reearea}[2]{\multicolumn{4}{@{}l}{\textbf{#1} (#2)}\\[1pt]}
\begin{table}[t]
\caption{REE covers 23 domains and 91 sub-fields in five areas. For each domain, the table gives its number of tasks, example sub-fields, and the main data modalities of its tasks.}
\label{tab:composition}
\centering
\scriptsize
\setlength{\tabcolsep}{4.3pt}
\renewcommand{\arraystretch}{1.0}
\begin{tabular}{@{}lrll@{}}
\toprule
Domain & Tasks & Sub-fields (examples) & Modalities \\
\midrule
\reearea{Mathematics \& Computing}{20}
Operations research \& planning & 6 & vehicle routing, crane scheduling, motion planning & mathematical instances, geometry \\
Algorithms \& discrete mathematics & 4 & sorting networks, circuit minimization & mathematical instances \\
Statistical learning \& forecasting & 4 & weakly supervised learning, change points & tables, graphs \\
Geometry \& graph drawing & 2 & graph drawing, computational geometry & geometry, graphs \\
Automated reasoning \& games & 2 & term rewriting, game-tree search & mathematical instances \\
Continuous \& black-box optimization & 2 & black-box optimization & mathematical instances \\
\addlinespace[2pt]\reearea{Physical Sciences \& Engineering}{25}
Materials science & 9 & crystal stability, topological materials & molecules, tables \\
Acoustics & 6 & source separation, sound events & audio \\
Chemistry & 5 & ligand binding, vapor pressure & molecules, images \\
Engineering \& signal processing & 3 & design, mesh segmentation, compression & geometry, images \\
Physics \& astronomy & 2 & quantum error correction, detectors & signals, tables \\
\addlinespace[2pt]\reearea{Life \& Health Sciences}{19}
Molecular biology \& genomics & 9 & gene regulation, microscopy & tables, sequences \\
Clinical medicine & 5 & dementia progression, MRI & text, images \\
Neuroscience & 5 & motor decoding, brain age & signals, tables \\
\addlinespace[2pt]\reearea{Earth \& Environmental Sciences}{12}
Water, air \& climate & 6 & air quality, wind power & signals, tables \\
Solid earth \& geophysics & 4 & seismology, petrology & tables \\
Remote sensing \& planetary science & 2 & urban mapping, wildfire & geometry, tables \\
\addlinespace[2pt]\reearea{Language, Society \& Culture}{24}
Linguistics & 8 & Latin morphology, phonetics & text, audio \\
Science of science & 4 & research impact, scholarly networks & graphs, text \\
Information \& knowledge systems & 4 & fact checking, question answering & text \\
Stylometry \& media studies & 4 & authorship, stance & text \\
Economics, finance \& education & 2 & finance, urban mobility & graphs, text \\
Music & 2 & harmony, musical key & audio \\
\midrule
\textbf{Total} & \textbf{100} & \textbf{91 sub-fields} & \\
\bottomrule
\end{tabular}
\end{table}}

For each task, REE releases the problem statement, the starter solution, the development data, the held-out test instances, the grader, and the expert solution, together with the metric value of the starter solution and the upper bound.

\section{Experimental Setup}
\label{sec:evaluation}
\label{sec:protocol}

We evaluate four frontier agents, each a language model running in an agent harness. Claude Opus 5.5 runs in Claude Code and GPT-6.1 Sol in Codex, each in the harness of its developer, and GLM-5.3 and Grok 4.7 run in Claude Code. All four use their highest reasoning effort and maximum context window and decide themselves when to stop. A run ends when the agent stops or after 12 hours, and its final solution is evaluated. Each score comes from one run per agent and task.

For task $i$ of the $N = 100$ tasks, let $B_i$ be an agent's score and $E_i$ the expert baseline (\Cref{sec:scale}). We report the mean score $\bar B$, the mean margin $\bar M$, the win rate $W$, and $W_5$, the share of tasks with a margin above five points:
\begin{equation}
  \bar B = \frac{1}{N}\sum_{i=1}^{N} B_i, \qquad
  \bar M = \frac{1}{N}\sum_{i=1}^{N} \left(B_i - E_i\right), \qquad
  W = \frac{100}{N}\sum_{i=1}^{N} \mathbf{1}\!\left[B_i > E_i\right], \qquad
  W_5 = \frac{100}{N}\sum_{i=1}^{N} \mathbf{1}\!\left[B_i - E_i > 5\right].
  \label{eq:metrics}
\end{equation}
The means $\bar B$ and $\bar M$ come with 95\% bootstrap intervals that resample tasks, and every comparison between two solutions is paired task by task. We also report these statistics for groups of tasks, and means over areas, domains, and other groups are point estimates. Expert solutions are of five kinds (\Cref{app:anchor-rule}), and not all were developed for the problem itself. To check that the comparison does not depend on this, we also report $\bar M$ and $W$ on the tasks whose expert solution is an entry in the problem's organized evaluation or a method published for the problem.

To follow a run over time, we use five checkpoints: its intermediate solutions evaluated at 1, 2, 4, and 8 hours (\Cref{sec:day}) and its final solution. Only runs evaluated at all five checkpoints enter the measures by checkpoint. When such a measure compares two agents, as the margin by hour and the share of the final score do, it is computed over the tasks on which both agents have such a run; otherwise it is computed over each agent's own runs.

These statistics show how well the agents perform, and a case study then examines how the results come about. It follows Claude Opus 5.5 and GPT-6.1 Sol, the two leading agents on every statistic of \Cref{eq:metrics}, through three records of their runs: the saved versions, the experiment logs, and the final methods. The saved versions and the logs give measures of how a run proceeds, such as its length and its number of versions, which \Cref{app:measures} defines. The final methods are analyzed by a judge based on a large language model (LLM), which describes each method and traces the central idea of the best final method of the two leading agents on each task through papers, write-ups, and public code (\Cref{app:audit}).

\section{Results}
\label{sec:results}

We organize the results around three questions: how well current research agents perform on these problems (\Cref{sec:performance}), how they compare with the expert solutions (\Cref{sec:main-results}), and how the leading agents arrive at their results (\Cref{sec:analysis}). \Cref{tab:main} gives the main numbers for the first two, and a case study of the two leading agents addresses the third.

\begin{table}[t]
\caption{Claude Opus 5.5 and GPT-6.1 Sol lead the four agents and score above the expert baseline. Each agent is a model in its harness (\Cref{sec:protocol}), with Claude Opus 5.5, GLM-5.3, and Grok 4.7 in Claude Code and GPT-6.1 Sol in Codex. Agents are ordered by mean score $\bar B$ (0 = the starter solution, 100 = the upper bound). $\bar M$ is the mean margin over the expert baseline, $W$ is the win rate, and $W_5$ is the share of tasks with a margin above five points (\Cref{eq:metrics}). Brackets give 95\% bootstrap intervals over tasks, and margins are computed before rounding. The expert baseline is its own row. Shading marks the top three values in each column, from darkest (best, bold) to lightest.}
\label{tab:main}
\centering
\footnotesize
{\newcommand{\rkone}[1]{\cellcolor{twblue!24}{\bfseries #1}}\newcommand{\rktwo}[1]{\cellcolor{twblue!13}#1}\newcommand{\rkthree}[1]{\cellcolor{twblue!5}#1}
\setlength{\tabcolsep}{7pt}
\begin{tabular}{@{}lcccc@{}}
\toprule
Agent & Score $\bar B$ [95\% CI] & Margin $\bar M$ [95\% CI] & Win rate $W$ (\%) & $W_5$ (\%) \\
\midrule
Claude Opus 5.5 & \rkone{39.6 \,[36.0, 43.2]} & \rkone{+15.2 \,[12.0, 18.3]} & \rkone{86} & \rkone{72} \\
GPT-6.1 Sol & \rktwo{38.2 \,[34.2, 42.3]} & \rktwo{+13.8 \,[10.3, 17.3]} & \rktwo{81} & \rktwo{69} \\
GLM-5.3 & \rkthree{28.6 \,[25.0, 32.3]} & \rkthree{+4.2 \,[0.8, 7.5]} & \rkthree{67} & \rkthree{56} \\
Grok 4.7 & 25.3 \,[21.9, 28.6] & +0.9 \,[$-$3.1, 4.7] & 61 & 44 \\
\midrule
Expert baseline & 24.4 \,[20.7, 28.4] & -- & -- & -- \\
\bottomrule
\end{tabular}}
\end{table}

\subsection{Agent Performance}
\label{sec:performance}

\begin{figure}[t]
\centering
\includegraphics[width=\linewidth]{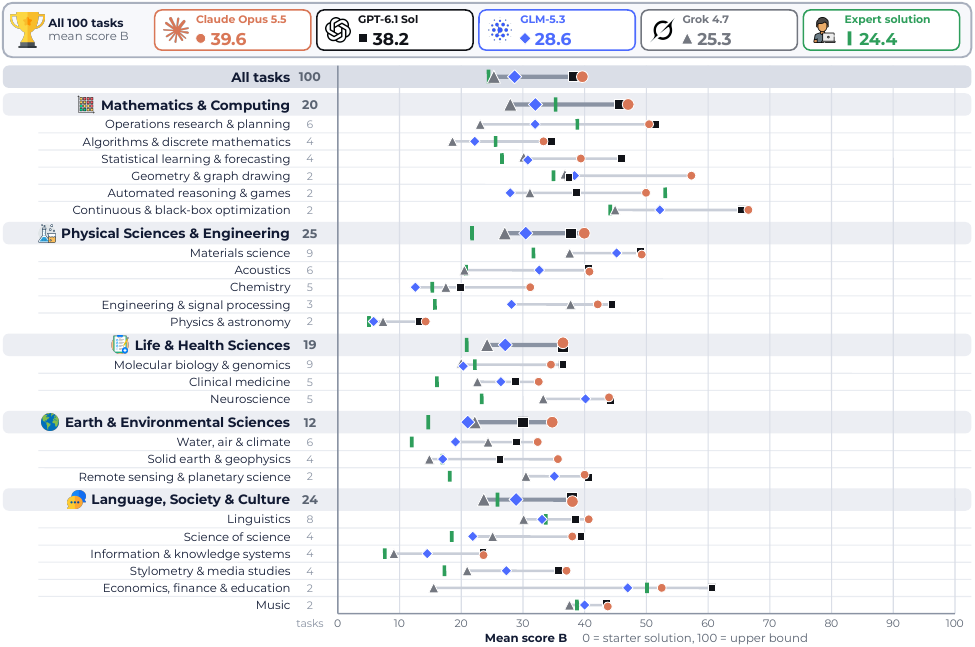}
\caption{Claude Opus 5.5 and GPT-6.1 Sol are the two highest-scoring agents in every area and improve on the starter solution in all 23 domains. Each row gives the mean scores $B$ of the four agents and of the expert solutions, on the scale from the starter solution (0) to the upper bound (100). The rows cover all tasks, each area (shaded), and each domain, with the number of tasks beside each name. A grey line spans the agents of a row.}
\label{fig:domains}
\end{figure}

\paragraph{Score.}
\label{sec:gap}
Claude Opus 5.5 and GPT-6.1 Sol reach mean scores of 39.6 and 38.2, and GLM-5.3 and Grok 4.7 reach 28.6 and 25.3 (\Cref{tab:main}). For the two leading agents the improvement is broad. Their mean score is above the starter solution in all 23 domains, with lowest domain means of 14.2 and 13.2 (\Cref{fig:domains}), and Claude Opus 5.5 improves on the starter solution on every task and GPT-6.1 Sol on all tasks but two. The scores nevertheless remain far from the upper bound. No run reaches the upper bound of its task, and the highest final score on any task is 81.7, which leaves room to measure stronger agents.

\begin{figure}[t]
\centering
\includegraphics[width=\linewidth]{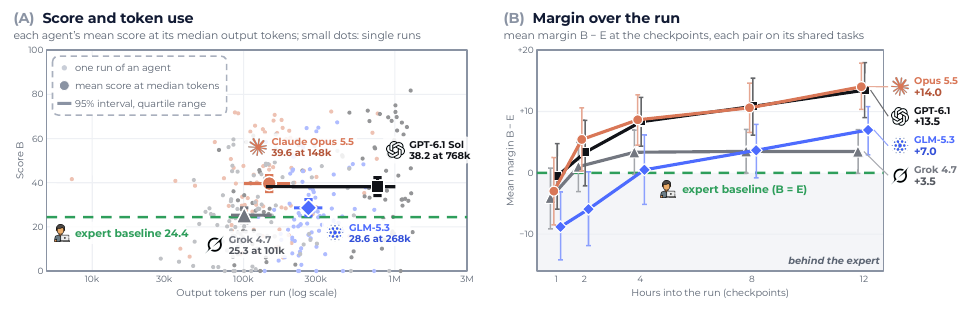}
\caption{Claude Opus 5.5 and GPT-6.1 Sol reach similar scores with about a fivefold difference in output tokens per run (a), and their mean margins are positive by hour 2, with intervals clear of zero by hour 4 (b). (a) Each agent's mean score at its median output tokens, with its 95\% bootstrap interval over tasks and the interquartile range of output tokens. Small dots are single runs with token records, each counted by its own model's tokenizer, and the dashed line is the mean expert baseline. (b) Each agent's mean margin $B - E$ at the five checkpoints, with 95\% bootstrap intervals over tasks, each pair of agents on the tasks where both runs are evaluated at every checkpoint. The dashed line is the expert baseline ($B = E$), and the 12-hour points are the final margins on these tasks, which \Cref{tab:main} gives over all tasks.}
\label{fig:tokencurves}
\end{figure}

\paragraph{Efficiency.}
\label{sec:tokens}
The two leading agents reach similar scores with very different token use (\Cref{fig:tokencurves}(a)). The median run of GPT-6.1 Sol uses about five times as many tokens as the median run of Claude Opus 5.5, in output tokens and in total tokens alike, with each model's tokens counted by its own tokenizer (\Cref{tab:exam} in \Cref{app:groups}). For each agent, the score of a run is only weakly correlated with its output tokens, with Spearman correlations between 0.20 and 0.31.

\subsection{Comparison with Expert Solutions}
\label{sec:main-results}

\paragraph{Margin.} The expert solutions reach a mean score of 24.4 (\Cref{tab:main}). Claude Opus 5.5 and GPT-6.1 Sol exceed this expert baseline by mean margins of 15.2 and 13.8 points, and both intervals exclude zero. For both agents the margin is positive in each of the five areas, where it lies between 10.4 and 20.1 points (\Cref{tab:groups} in \Cref{app:groups}). It also holds on the tasks whose expert solution was developed for the problem itself, with margins of 12.2 points [8.7, 15.9] and 12.3 points [8.2, 16.4]. The other two agents, in contrast, are close to the expert baseline, with mean margins of 4.2 points for GLM-5.3 and 0.9 for Grok 4.7, whose interval includes zero.

\begin{figure}[t]
\centering
\includegraphics[width=\linewidth]{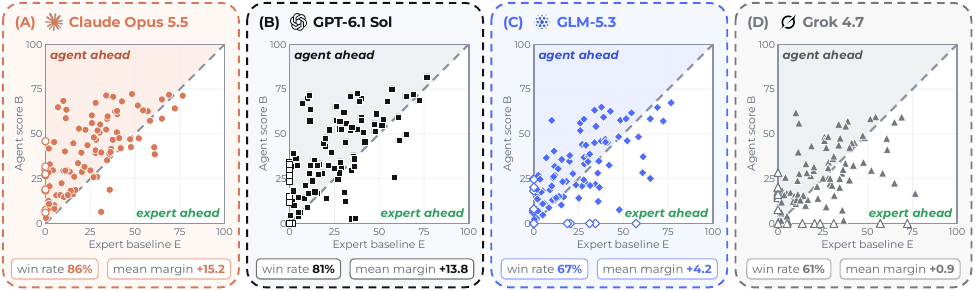}
\caption{Task by task, Claude Opus 5.5 and GPT-6.1 Sol score above the expert baseline on 86 and 81 of the 100 tasks. Each panel shows one agent, and each point is one task, with the agent's score $B$ against the expert baseline $E$. Above the dashed line the agent's final solution scores higher than the expert solution. Open markers are tasks where either score is clipped at 0, at or below the starter solution. Each panel also gives the agent's win rate and mean margin.}
\label{fig:parity}
\end{figure}

\paragraph{Win rate.} The win rate shows how widely this advantage holds across tasks. Claude Opus 5.5 and GPT-6.1 Sol score above the expert baseline on 86\% and 81\% of tasks, and by more than five points on 72\% and 69\% (\Cref{tab:main}). As with the margin, the result holds across groups of tasks: the win rate of both agents is at least 67\% in each of the five areas, and it is 83\% and 82\% on the tasks whose expert solution was developed for the problem itself. The other two agents win less often: GLM-5.3 and Grok 4.7 score above the expert baseline on 67\% and 61\% of tasks, and by more than five points on 56\% and 44\%. \Cref{fig:parity} shows the same contrast, with the points of the two leading agents mostly above the dashed line and those of the other two spread more evenly around it.

\subsection{Case Study}
\label{sec:stages}
\label{sec:analysis}

To see how the leading agents arrive at these results, we follow Claude Opus 5.5 and GPT-6.1 Sol through their runs. We begin with one run in detail and then examine, over all of their runs, how progress develops over time, how the methods are built, and how the agents validate and select their solutions.

\subsubsection{A Run in Detail}
\label{sec:inside}

\begin{figure}[t]
\centering
\includegraphics[width=\linewidth]{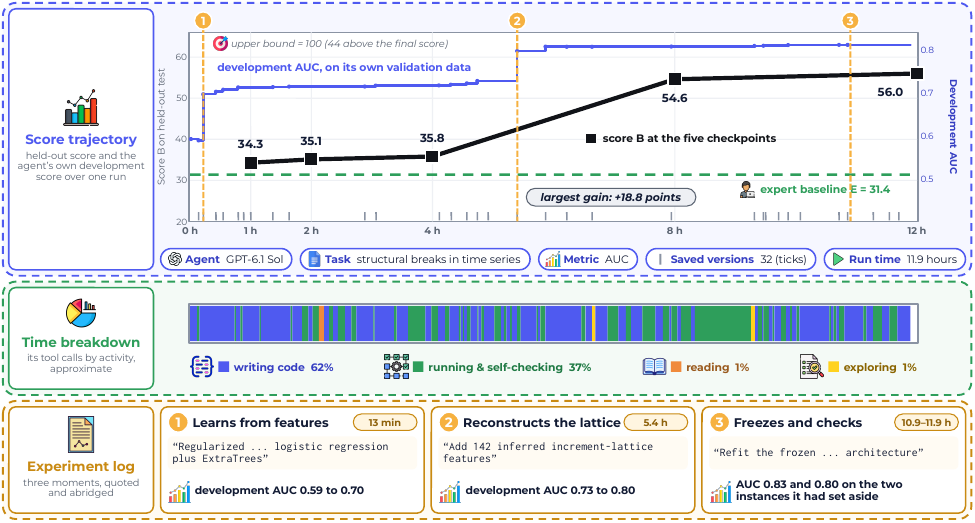}
\caption{GPT-6.1 Sol surpasses the expert solution by its first checkpoint on structural breaks in time series, and its largest gain, from 35.8 at four hours to 54.6 at eight, comes after the lattice features that it added at 5.4 hours. \emph{Top}: the score $B$ at the five checkpoints, the expert baseline, and the development AUC after each measured version (the task's metric on the agent's own validation data, which is not $B$), with ticks for the saved versions. \emph{Middle}: the agent's tool calls by activity over the run, approximate. \emph{Bottom}: three moments from its experiment log, quoted and abridged, which the numbers 1 to 3 mark on the time axis.}
\label{fig:insiderun}
\end{figure}

The run is that of GPT-6.1 Sol on structural breaks in time series. \Cref{fig:insiderun} follows it from its saved versions and its experiment log, and \Cref{app:run-full} gives it in full. The task asks the agent to rank time-series records by the evidence that their generating rules changed at a given boundary, where the change may concern the variability or the temporal dependence of a series as well as its mean. Each instance is scored by the area under the ROC curve (AUC), and the starter solution ranks each record by the relative difference between the variances of its two segments, using no labels. The development scores quoted below are the task's AUC on the agent's own validation data, not scores on the REE scale.

Before writing any code, the agent split its development data. It set some instances aside as a final check, writing that their labels ``will not be inspected until a candidate is selected'', and validated on the rest by leaving out one instance at a time. At 13 minutes it replaced the rule of the starter solution with a learned blend of a regularized logistic regression and extremely randomized trees, which raised its development score from 0.59 to 0.70. The version current at one hour scored 34.3 on the held-out test instances, already above the expert baseline.

Over the next hours the agent tried many variants, including likelihood features from fitted volatility models, kernel machines trained partly on simulated series, boosted trees, and rank fusion. Its development score rose slowly while its score at the checkpoints changed little, reaching 35.8 at four hours. Late in this period the agent noticed that some series take only a limited set of values, and it added features for repeated values and decimal resolution. At 5.4 hours it took this observation further: it reconstructed a multiplicative path behind the observed values and added 142 features from the inferred lattice of increments on either side of the boundary. Its development score rose from 0.73 to 0.80, a gain that it recorded as ``large \ldots{} across all instances'', and the version current at eight hours scored 54.6.

From 10.9 hours the agent froze the method and refit it on all the development instances that it had validated on. Only then did it score the instances that it had set aside, a single time. They gave 0.83 and 0.80, in line with its development estimate of 0.81, and the agent kept this version as its final solution. It saved 32 versions and ended the run at 11.9 hours with a final score of 56.0, the highest of any run on the task and about 44 points below the upper bound.

Three features of this run return in the analyses below. The agent kept improving late, making its largest gain after the four-hour checkpoint and using nearly all of its 12 hours (\Cref{sec:curves}). That gain came from a pattern that it had found in the data (\Cref{sec:sources}). Finally, it kept a check that its tuning never touched (\Cref{sec:checks}).

\subsubsection{Progress over Time}
\label{sec:curves}

On average, the leading agents surpass the expert solutions early in a run. Their mean margins are $-$3.0 and $-$0.5 points after one hour and +5.4 and +3.4 after two (\Cref{fig:tokencurves}(b)). At two hours the 95\% interval of Claude Opus 5.5 lies above zero (lower bound +0.3), and that of GPT-6.1 Sol does not yet ($-$1.9). By four hours, at +8.6 and +8.3, both intervals lie above zero (lower bounds +4.7 and +4.3), and the mean scores of both agents reach 87\% of their mean final scores. GLM-5.3, by comparison, starts below the expert baseline and has a margin of +0.5 points by four hours.

\begin{figure}[t]
\centering
\includegraphics[width=\linewidth]{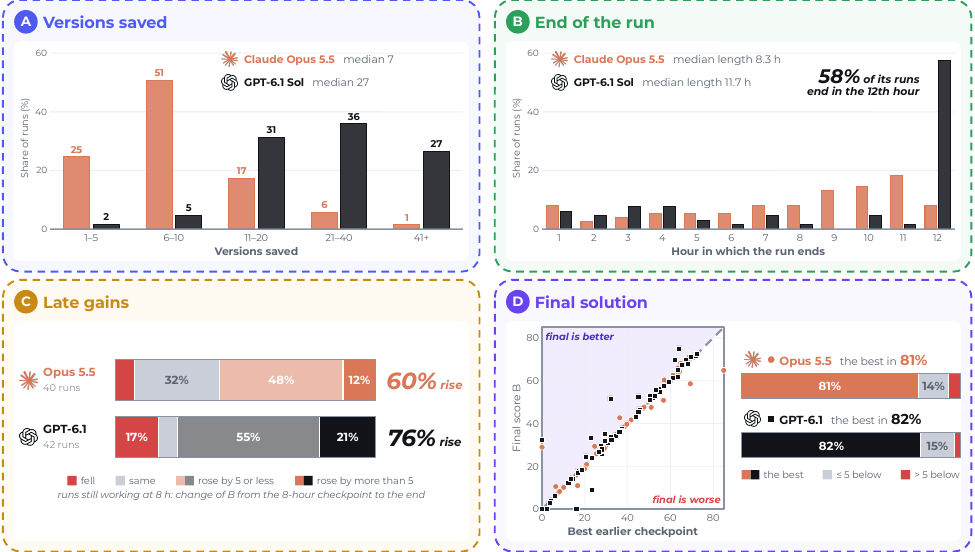}
\caption{The two leading agents use a run differently. GPT-6.1 Sol saves many more versions per run than Claude Opus 5.5, and most of its runs end in the twelfth hour, while the runs of Claude Opus 5.5 end throughout the 12 hours. For both, the final solution is usually the best of the evaluated ones, and runs that are still working at eight hours often still improve. (a) Share of runs by the number of versions saved. (b) Share of runs ending in each hour. (c) Change of the score from the eight-hour checkpoint to the end, over the runs still working at eight hours. (d) The final score against the best earlier checkpoint, and the share of runs in which the final solution is the best of the evaluated ones, over the runs evaluated at every checkpoint.}
\label{fig:runuse}
\end{figure}

Progress does not end there, and runs that keep working often still improve late. Gains after four hours can be large, as in the run above, which made its largest gain between its four-hour and eight-hour checkpoints. Among the runs that are still working at eight hours, the score rises between the eight-hour checkpoint and the end in 60\% of the runs of Claude Opus 5.5 and 76\% of the runs of GPT-6.1 Sol, and by more than five points in 12\% and 21\% (\Cref{fig:runuse}(c)).

Many runs nevertheless end well before the 12 hours are over, and the agents differ in how much of the time they use. Claude Opus 5.5 ends 74\% of its runs with at least two hours left, and GPT-6.1 Sol 39\%. The runs of Claude Opus 5.5 end throughout the 12 hours, while most runs of GPT-6.1 Sol end in the last hour (\Cref{fig:runuse}(b)), and their median runs stop after 8.3 and 11.7 hours, which leaves 3.7 and 0.3 hours unused. On average, Claude Opus 5.5, GPT-6.1 Sol, and GLM-5.3 run for 7.3, 8.6, and 7.9 hours, and 30\%, 30\%, and 26\% of their runs end within six hours. Grok 4.7 stops after 2.6 hours on average and ends 91\% of its runs within six hours, in part because of how it interacts with its harness (\Cref{app:runs}). Within each agent, run length and score are only weakly associated, with Spearman correlations of 0.05 for Claude Opus 5.5 and 0.31 for GPT-6.1 Sol.

The two leading agents also pace their work differently. Claude Opus 5.5 saves a median of 7 versions per run and GPT-6.1 Sol 27. Most runs of Claude Opus 5.5 save ten versions or fewer, while most runs of GPT-6.1 Sol save more than ten (\Cref{fig:runuse}(a)). GPT-6.1 Sol thus works in many small steps, and Claude Opus 5.5 in fewer, larger ones.

\subsubsection{How the Methods Are Built}
\label{sec:sources}

The best methods of the leading agents are mostly built from methods that their fields already have, and what the agents contribute is how they tailor and combine them. For each task, an LLM judge traces the best final method of the two leading agents to prior work (\Cref{app:audit}). On 51\% of tasks it is the standard method of the field, tailored closely to the data and the metric of the problem. On 45\% it combines existing methods, often from different strands of a field, into a pipeline built for the problem, and the judge rates 96\% of these combinations as non-obvious. One method is transferred from another field, and three are left unclassified. Counting the non-obvious combinations and the transferred method, the best method goes beyond tailoring on 44\% of tasks. The examples below follow the sources of these methods outward, from the data of the task to the methods of the field and to other fields.

The nearest source is the data of the task. What an agent observes there often shapes its method, and such an observation can come hours into a run, as the lattice of increments did in the run above. In seizure detection from EEG, Claude Opus 5.5 finds that models trained across patients fail on its validation data, drops them, and judges each recording against its own range, which lifts its development score from 0.14 to 0.53. On the held-out test instances the run scores 33.0, the highest of the four runs on the task. On structural breaks in time series, both leading agents notice that some series take a limited set of repeated values.

\begin{table}[t]
\caption{On these six tasks, the leading agents use a method of their own in place of the method of the expert solution. Each row gives the expert solution, what the two leading agents do instead, and their scores. $^{\times}$ marks a run below the expert baseline.}
\label{tab:methods}
\centering
\scriptsize
\setlength{\tabcolsep}{3pt}
\begin{tabular}{@{}>{\raggedright\arraybackslash}p{0.2\linewidth}>{\raggedright\arraybackslash}p{0.235\linewidth}>{\raggedright\arraybackslash}p{0.37\linewidth}cc@{}}
\toprule
Problem & Expert solution & What the agents do differently & \shortstack{Claude\\Opus 5.5} & \shortstack{GPT-6.1\\Sol} \\
\midrule
CO\textsubscript{2} uptake of porous frameworks & Random forest on published descriptors & Force-field CO\textsubscript{2} energies inside the pores & 60.3 & 65.2 \\
Commuting flows between regions & Graph-attention embeddings of origins and destinations & A gravity model of spatial interaction stacked with flow features in one run, forests and a neural mean model in the other & 40.8 & 52.8 \\
News stance detection & Small network on term counts & Separate relatedness and stance stages, with refutation and hedging features & 51.9 & 52.7 \\
Structural breaks in time series & Boosted trees on 78 features & Wider boundary features, with an inferred lattice of increments in one run & 40.1 & 56.0 \\
Gene trees from protein sequences & Alignment-free word statistics & Alignment and tree search, with a likelihood model in one run & 47.0$^{\times}$ & 55.5 \\
Confluence proofs for rewriting systems & Prover portfolio written by experts & Their own prover portfolio, with one run reporting testing on about 800 instances of its own & 57.3$^{\times}$ & 25.8$^{\times}$ \\
\bottomrule
\end{tabular}
\end{table}

The agents also look beyond the method of the expert solution, and on some tasks they set it aside (\Cref{tab:methods}). On CO\textsubscript{2} uptake, both compute the force-field and electrostatic energies of the CO\textsubscript{2} molecule inside the pores. On commuting flows, Claude Opus 5.5 fits the gravity model of spatial interaction to each instance before learning the rest, and on gene trees both align the protein sequences with substitution matrices.

Other strong methods combine an existing method of the field with components that the agent adds. In inferring gene regulatory networks from single-cell snapshots, GPT-6.1 Sol takes directional evidence from how the distribution of expression changes along the observation order, as SINCERITIES does \citep{papili2018sincerities}, adds a prior over cyclic graphs, and trains an edge scorer on thousands of networks drawn from a simulator. With the trained scorer, the run reaches 71.7, the highest score on the task. In predicting chlorine concentrations in a drinking-water network, Claude Opus 5.5 identifies the hydraulics of each scenario with the Hazen--Williams relation, moves chlorine through the pipes with a Lagrangian transport simulator of the kind that EPANET implements \citep{rossman2000epanet}, and takes the median prediction over an ensemble of pipe volumes. The run reaches 67.3, ahead of every other run on the task.

The farthest source is another field. On shifted and rotated test functions whose family structure the statement makes public, Claude Opus 5.5 brings a technique from signal processing into black-box optimization. It locates the hidden shift of a Schwefel component with a matched filter, the technique used to find a known pulse in a noisy signal. This step cuts its development error from 1.47 to 0.53, and the run has the highest score on the task at 72.2.

\subsubsection{Validation and Selection}
\label{sec:checks}

The leading agents choose well among their own solutions in most runs, and how far their checks can be trusted depends on the data that the checks use. Because an agent receives no score on the held-out test instances (\Cref{sec:day}), it chooses its final solution with checks of its own design. In most runs this choice is the best one available: the final solution scores at least as high as every earlier checkpoint in 81\% of the runs of Claude Opus 5.5 and 82\% of the runs of GPT-6.1 Sol (\Cref{fig:runuse}(d)). In the remaining runs an earlier version would have scored higher, by more than five points in 6\% and 3\% of runs.

A check on data that the agent's tuning never touched shows whether a development estimate holds. In the run above, the instances that GPT-6.1 Sol had set aside confirmed its development estimate. When the development data stop separating candidates, an agent can also build such data itself. On confluence proofs, both leading agents certify 8 of 9 development systems, which then no longer separate their versions. Claude Opus 5.5 goes on to test on instances that it writes or generates itself, about 800 by its own log, and its final prover certifies 62 of 100 held-out systems. That of GPT-6.1 Sol certifies 38 and exceeds the time limit on 4 more, which the task penalizes, and both score below the expert solution (\Cref{tab:methods}).

Conversely, a check gives little information when it uses the data that the method was fitted on. In identifying verbal multiword expressions, GPT-6.1 Sol checks its classifier on validation folds, but it builds its phrase list from all the development answers and selects its last refinements on the full development data. Its check on the full development data gives an F1 of 0.996, against 0.278 for the starter solution on the same data, while its solution reaches an F1 of 0.288 on the held-out test instances, below the 0.319 of the starter solution. On the same task, Claude Opus 5.5 trains on the same answers and adds its own lexical knowledge, but it also keeps a check on data that it does not train on. Its last check on all the development instances gives an F1 of about 0.81, and its solution reaches 0.512 on the held-out test instances.

In summary, the case study shows three things about how the leading agents work. They surpass the expert solutions early in a run and often still improve when they keep working. Their best methods mostly tailor or combine the existing methods of their fields. Their own validation selects their best solution in most runs, and it is most informative when it uses data kept apart from fitting.

\section{Limitations}
\label{sec:limitations}

REE covers research problems whose solutions a program can score on held-out instances within a fixed time limit, so research that needs physical experiments, new data collection, or human judgment of a result lies outside its scope. Each score comes from a single run, and the bootstrap intervals resample tasks, so they describe the spread of scores over tasks and do not describe the variation between repeated runs on one task. Each agent also runs under one budget and in one harness, with 12 hours of wall-clock time at its highest reasoning effort, and other budgets or harnesses may give other scores.

For most tasks the upper bound is the best value that the metric can take by definition, so the distance to it measures room on the scale, which a method may not be able to close. Each task has one expert solution, evaluated within the budget of the task, and its score can differ from the performance reported in its original publication. The tracing of methods to prior work relies on an LLM judge.

\section{Conclusion}
\label{sec:conclusion}

We have introduced REE, a benchmark of 100 open-ended research problems in five scientific areas and 23 domains. Each problem asks an agent to iteratively improve a starter solution within 12 hours, and REE scores the result on one scale together with an expert solution evaluated under comparable conditions. On this benchmark, the two leading agents outperform the expert solutions on 86\% and 81\% of problems, with mean margins of 15.2 and 13.8 points, and their mean score is above the expert baseline in every area. They surpass the expert solutions within two hours on average, and their best solutions mostly tailor or combine existing methods of their fields. With mean scores of 39.6 and 38.2 out of 100, no agent approaches the upper bound, so the benchmark can continue to measure stronger agents. Because the scale and the expert solutions are fixed, later agents can be compared on the same problems, and we release the problem statements, starter solutions, development data, held-out test instances, graders, and expert solutions to support this.

\bibliographystyle{assets/plainnat}
\bibliography{references}

\clearpage
\beginappendix

\section{Kinds of Expert Solution}
\label{app:anchor-rule}

The expert solution of a task is an established human solution of one of five kinds.
\begin{enumerate}
  \item An entry in the problem's organized evaluation, in which entries to one problem are ranked on a common metric (24 tasks).
  \item A method published for the problem, in a peer-reviewed paper or its authors' code (36 tasks).
  \item A reference method released with the problem's benchmark, dataset or toolkit (20 tasks).
  \item A released tool for the problem, with evidence of standing: wide use, use in later papers, or adoption as an official reference method (9 tasks).
  \item A standard method of the field that differs from the starter solution (11 tasks).
\end{enumerate}
\Cref{sec:main-results} gives the margin and the win rate on the first two kinds, the solutions built for the problem itself.

\section{What Makes the Problems Challenging}
\label{app:hard}

The problems are challenging for reasons that go beyond implementation effort, and five situations, each shown on one task, illustrate the decisions that an agent must make with the data it has.

\paragraph{Unseen sources.} On some tasks the held-out test instances come from sources that the development data do not contain. In seizure detection from EEG, each test recording comes from a source of its own that does not appear in the development data, so a method tuned to the patients that the agent can see may not hold on new ones.

\paragraph{The field's own metric.} Each task keeps the metric that its field uses, which can reward something that a generic loss does not. In building gene trees from protein sequences, the solution outputs distances that a fixed neighbor-joining procedure turns into a tree, and the score counts the branchings that differ from a curated tree, ignoring the root and the branch lengths, so distances that are accurate on average can still yield a wrong tree.

\paragraph{Approximate models.} The standard model of a problem can be an approximation that the data violate. In decoding quantum error-correction experiments, the supplied circuit model approximates the noise as Pauli errors, while the measured shots from superconducting qubits can also contain leakage and correlated faults outside that model, which a decoder built on the model alone does not account for.

\paragraph{Strong expert solutions.} Some fields have mature methods that encode years of expert work. In confluence proofs for term rewriting systems, a property that is undecidable in general, a claim counts only when an independent proof verifier certifies it. The expert solution is a portfolio of provers written by experts in the field, so an agent that surpasses it must build a prover of comparable strength within 12 hours.

\paragraph{A fixed budget.} Each solution must finish within the time limit that its statement declares, and the agent has 12 hours to build it. In mate search in chess, only the exact distance to mate earns credit for a position and the method is stopped at the deadline of each instance, so the agent must trade the depth of its search against the time it has.

\section{Run Settings}
\label{app:runs}

Each run gets four CPUs, 16\,GB of memory and no network access beyond its model interface, for 12 hours of wall time. The inference of the agent's model lies outside this compute budget, and \Cref{tab:exam} reports its token use. The context window is set to each model's maximum. Every version of the solution that the agent saves is stored with its timestamp and the tokens spent so far. The final solution is evaluated in a separate container with four CPUs and 8\,GB of memory within the time limit that the statement declares, and the starter solution and the expert solution are evaluated by the same grader with the same memory and time.

Grok 4.7 runs in Claude Code rather than in a harness of its developer's, and the pairing costs it. In this harness Grok 4.7 often closes its turn with a reply in text alone that announces further work, such as the next test it will run, without the tool call that would start it. Claude Code, run non-interactively, takes a turn without a tool call as the end of the task, so such a run stops with hours left. GLM-5.3, in the same harness, does not end its runs this way. Grok 4.7's scores are therefore those of Grok 4.7 in Claude Code, and a harness built for the model may take it further.

\section{Tracing Strong Methods}
\label{app:audit}

For each task, an LLM judge, a separate session of GPT-6 Astra at its highest reasoning effort and with web search, reads the four runs beside the expert solution, including the trajectory, final method, experiment log, and scores of each run. It describes each final method and the components that give its gain, and it traces the central idea of the best final method of the two leading agents through papers, write-ups and public code, searching for each named component and for public implementations of it. It then places that method in one of four classes. The classes are existing methods combined into a pipeline built for the problem, the standard method of the field tailored to the data and the metric of the problem, a method transferred from another field and adapted, and an idea with no close prior work. When the evidence cannot settle a class, it leaves the method unclassified. Combinations it judges non-obvious, and methods transferred across fields, count as going beyond tailoring in \cref{sec:sources}.

\section{Results by Group}
\label{app:groups}

\Cref{tab:groups} breaks the main result down by area, problem type, and data modality, and \Cref{tab:exam} gives the token use behind \Cref{fig:tokencurves}(a).

\begin{table}[!ht]
\caption{The two leading agents have a positive mean margin in every area. Each cell gives an agent's mean margin $B - E$ by area, problem type, and data modality, and in parentheses its win rate in percent. Types and modalities with fewer than five tasks share one row, and modalities use the short names of \Cref{tab:composition} (\Cref{fig:composition} gives the full names).}
\label{tab:groups}
\centering
\scriptsize
\begin{tabular}{@{}lcccc@{}}
\toprule
 & Claude Opus 5.5 & GPT-6.1 Sol & GLM-5.3 & Grok 4.7 \\
\midrule
\multicolumn{5}{@{}l}{\textbf{Area}} \\
Earth \& Environmental Sciences & +20.1 (100) & +15.4 (92) & +6.4 (75) & +7.5 (92) \\
Physical Sciences \& Engineering & +18.2 (96) & +16.0 (88) & +8.7 (80) & +5.3 (68) \\
Life \& Health Sciences & +15.6 (79) & +15.6 (84) & +6.3 (79) & +3.3 (74) \\
Language, Society \& Culture & +12.2 (83) & +12.1 (67) & +3.0 (62) & $-$2.2 (54) \\
Mathematics \& Computing & +11.8 (75) & +10.4 (80) & $-$3.3 (40) & $-$7.3 (30) \\
\midrule
\multicolumn{5}{@{}l}{\textbf{Problem type}} \\
Regression & +21.4 (95) & +16.6 (86) & +8.4 (77) & +9.0 (82) \\
Inverse problems & +17.9 (88) & +22.7 (88) & +4.0 (75) & +5.9 (75) \\
Structured prediction & +16.6 (72) & +15.2 (67) & +5.2 (67) & $-$2.6 (61) \\
Classification & +13.3 (100) & +12.6 (87) & +6.7 (78) & +1.8 (65) \\
Combinatorial optimization & +10.0 (64) & +6.3 (71) & $-$7.0 (21) & $-$11.6 (14) \\
Ranking \& embedding & +5.0 (67) & +5.0 (67) & $-$4.0 (33) & $-$3.4 (33) \\
Forecasting, continuous opt.\ and modeling & +14.8 (100) & +16.8 (100) & +8.8 (100) & +3.5 (78) \\
\midrule
\multicolumn{5}{@{}l}{\textbf{Data modality}} \\
Signals & +20.4 (100) & +14.9 (92) & +6.3 (77) & +6.9 (77) \\
Molecules & +18.2 (100) & +18.0 (92) & +12.3 (92) & +8.1 (75) \\
Audio & +15.5 (89) & +15.5 (78) & +8.9 (67) & $-$0.5 (44) \\
Graphs & +14.7 (100) & +12.6 (80) & +7.8 (100) & +5.5 (100) \\
Text & +14.4 (76) & +13.0 (67) & +3.7 (57) & $-$1.8 (52) \\
Tables & +12.4 (94) & +11.9 (81) & +1.7 (69) & +0.4 (75) \\
Mathematical instances & +10.2 (64) & +8.9 (79) & $-$1.8 (36) & $-$6.8 (21) \\
Geometry, images and sequences & +17.9 (80) & +18.0 (90) & $-$0.6 (70) & +0.6 (70) \\
\bottomrule
\end{tabular}
\end{table}

\begin{table}[!ht]
\caption{The two leading agents differ about fivefold in token use. The table gives each agent's median output and total tokens per run, over the runs with token records, and the Spearman correlation between a run's score and its output tokens. Total tokens are input tokens, cached ones included, plus output tokens.}
\label{tab:exam}
\centering
\footnotesize
\begin{tabular}{@{}lccc@{}}
\toprule
Agent & Output tokens & Total tokens & Spearman (score, output) \\
\midrule
Claude Opus 5.5 & 148\,k & 21.6\,M & 0.20 \\
GPT-6.1 Sol & 768\,k & 106.2\,M & 0.31 \\
GLM-5.3 & 268\,k & 85.9\,M & 0.22 \\
Grok 4.7 & 101\,k & 15.0\,M & 0.24 \\
\bottomrule
\end{tabular}
\end{table}

\clearpage
\section{Full Comparison with Related Benchmarks}
\label{app:related}

\Cref{tab:related-full} extends the comparison in \Cref{tab:related} with each benchmark's scoring rule and with ScienceAgentBench.

\paragraph{Scales and evaluation over time.} Benchmarks with continuous scores often define the scale by a reference. RE-Bench sets a reference solution to 1, MLRC-Bench sets the top competition entry to 100, and Agent4Sci reports the gain over the published best \citep{wijk2025rebench,zhang2025mlrcbench,agent4sci2026}. A scale defined this way moves when the reference improves. Other benchmarks fix both ends of the scale in advance, RSI-Exam between a starting solution and an upper bound and AI4AI-Bench between an uninformative model and the task optimum \citep{rsiexam2026,chi2026ai4aibench}. REE uses a scale of this kind, and the expert baseline is measured on it for every problem. Several benchmarks also follow a run over time. RE-Bench plots score against time for agents and experts \citep{wijk2025rebench}, PaperBench evaluates a run at several points \citep{starace2025paperbench}, LHTB~2.0 runs its verifier every 30 minutes \citep{li2026lhtb2}, and Agent4Sci plots running-best curves on validation data \citep{agent4sci2026}. In REE, intermediate solutions taken at fixed hours are evaluated on the held-out test instances.

\begin{table}[!ht]
\caption{The full comparison behind \Cref{tab:related}. \emph{Horizon}: wall-clock time of one agent run. \emph{Score}: C continuous, P/F pass/fail, R rubric, followed by what the ends of the scale stand for, where \emph{open} means that the scale has no fixed top. \emph{Human reference}: the human or expert result that agents are compared with, where \cmark{} means that it is evaluated under comparable conditions. \emph{Held-out}: the test instances are hidden from the agent, and it receives no score on them during the run. \emph{Over time}: solutions are also evaluated at intermediate times. \cmark{} yes, \pmark{} in part, \xmark{} no, -- not applicable, n.s.\ not stated in the source.}
\label{tab:related-full}
\centering
\scriptsize
\setlength{\tabcolsep}{2.5pt}
\renewcommand{\arraystretch}{1.12}
\begin{tabular}{@{}>{\raggedright\arraybackslash}p{0.125\linewidth} r >{\raggedright\arraybackslash}p{0.13\linewidth} l >{\raggedright\arraybackslash}p{0.17\linewidth} >{\raggedright\arraybackslash}p{0.17\linewidth} c c@{}}
\toprule
Benchmark & Tasks & Fields & Horizon & Score & Human reference & Held-out & Over time \\
\midrule
MLE-bench \citep{chan2025mlebench} & 75 & ML engineering on Kaggle problems from many fields, 15 problem types & 24\,h & P/F: Kaggle medal & \pmark{} Kaggle leaderboard, recorded & \pmark$^{a}$ & \pmark \\
RE-Bench \citep{wijk2025rebench} & 7 & ML R\&D & 8\,h$^{b}$ & C: 0 start, 1 author's reference; open & \cmark{} 71 runs by 61 experts, 8\,h, same VMs & \pmark$^{c}$ & \cmark \\
PaperBench \citep{starace2025paperbench} & 20 & ML, 12 ICML topics & 12\,h & R: 100\% = full replication & \pmark{} 8 ML PhDs on 4 papers & -- & \pmark \\
ScienceAgent\-Bench \citep{chen2025scienceagentbench} & 102 & 4 sciences & n.s. & P/F: success criteria & \xmark & \pmark & \xmark \\
MLRC-Bench \citep{zhang2025mlrcbench} & 7 & ML, conference competitions & 5--10\,h & C: 0 baseline, 100 top entry; open & \pmark{} top entry, recorded or re-run & \cmark & \pmark \\
ALE-Bench \citep{imajuku2025alebench} & 40 & Heuristic optimization & 4\,h$^{d}$ & C: contest performance rating; open & \pmark{} contestants, recorded & \cmark & \pmark \\
AIRS-Bench \citep{lupidi2026airsbench} & 20 & ML, 7 categories & 24\,h & C: 0 worst run, 1 literature SOTA; open & \pmark{} literature SOTA, quoted & \pmark$^{a}$ & \pmark \\
NatureBench \citep{wang2026naturebench} & 90 & 6 science domains & 4\,h & C: gap to paper SOTA; open & \pmark{} paper SOTA, quoted & \pmark$^{c}$ & \pmark \\
Agent4Sci \citep{agent4sci2026} & 25 & 11 science domains & 12\,h & C: gain over published best; open & \pmark{} published best, re-run as the start & \cmark & \pmark \\
FrontierCS~2 \citep{mang2026frontiercs2} & 48 & 13 computer-science domains & 5\,h & P/F: beats the reference on every metric; C: 0--100 per task & \cmark{} authors' code, same held-out suite & \cmark & \pmark \\
LHTB~2.0 \citep{li2026lhtb2} & 77 & 13 domains & 3\,h & C: progress ladder, 1 = all stages & \xmark & \cmark & \cmark \\
RSI-Exam \citep{rsiexam2026} & 88 & 6 domains & 12\,h & C: 0 start, 0.6 reference, 1 bound & \pmark{} frontier-calibrated reference run by the task author, 0.6 where one exists & \cmark & \pmark \\
\midrule
\textbf{REE (ours)} & \textbf{100} & \textbf{23 domains} & \textbf{12\,h} & \textbf{C: 0 starter, 100 upper bound} & \textbf{\cmark{} expert solution, comparable conditions} & \textbf{\cmark} & \textbf{\cmark} \\
\bottomrule
\end{tabular}

\smallskip
\raggedright\scriptsize
$^{a}$The agent sees the test inputs but not their answers.
$^{b}$Runs of 30\,min, 2\,h, and 8\,h, with best-of-$k$ totals up to 32\,h.
$^{c}$The agent sees its score on the test instances during the run.
$^{d}$Rules set the length of the original contest (about 4\,h to 2 weeks), and the paper's runs use 4\,h.
\end{table}

\clearpage
\section{The Recorded Run in Full}
\label{app:run-full}

The run of \Cref{sec:inside}, GPT-6.1 Sol on structural breaks in time series, saved 32 versions over 11.9 hours. Its experiment log records, for each version, the change, the measurements the agent took and its decision, and it records separately the branches the agent explored without changing its solution. Read in order, the run falls into four phases. Development scores are the task's AUC on the agent's own validation data.

\paragraph{Setting up (to 13 minutes).} The log opens with a plan for generalization, with validation by leaving one development instance out at a time and a set of instances set aside whose labels would not be inspected until a candidate was selected. The agent first computed a broad set of distribution, dependence and boundary-window features and tried an unsupervised combination of them, which it rejected while keeping the features. At 13 minutes it replaced the starter solution with a learned blend of a regularized logistic regression and extremely randomized trees, and its development score rose from 0.59 to 0.70.

\paragraph{Widening the method (to 5.4 hours).} Over the next versions the agent added likelihood features from fitted volatility models, kernel machines trained partly on series simulated from fitted models, a shallow boosted classifier and a rank fusion of the classifiers' outputs, judging each change on its leave-one-instance-out score and on a fixed validation split. Branches it explored and rejected included spectral features, neural networks, pairwise ranking losses and larger boosters. It also hardened the solution, capping extreme feature values and testing it on short, constant and extreme-scale records that it constructed itself. Shortly before the next phase it added features for repeated values and decimal resolution.

\paragraph{Reconstructing the lattice (5.4 to 10.9 hours).} The agent reconstructed a multiplicative path behind the observed values and added 142 features from the lattice of increments on either side of the boundary, with a simpler linear classifier on top, and its development score rose from 0.73 to 0.80. The following versions combined this classifier with a booster on the new features, routed records without a clear lattice to a separate classifier, and augmented the training data with sign-reversed and simulated records. The agent also bounded the cost of its feature extraction on very long records, so that the solution would finish within the time limit.

\paragraph{Freezing and checking (from 10.9 hours).} The agent froze the architecture, refit it on all the development instances it had validated on, removed unused models and files, and fixed two numerical edge cases, for extremely small values and for segments of a single sample. Only then did it score the instances that it had set aside, once. They gave 0.83 and 0.80, against a development estimate of 0.81. Its last check confirmed that its ranking did not depend on the order of the records. The final solution scored 56.0 on the held-out test instances.

\section{Measurement Definitions}
\label{app:measures}

The measures of \Cref{sec:stages} are defined below for each agent, grouped by the runs they are taken over. The last two paragraphs note how development scores are quoted and two special upper bounds (\Cref{sec:scale}).

\paragraph{Runs with recorded timings.} \emph{Run length} is the wall-clock time in hours from the start of the run to the agent's stop, and the mean run length and the share of runs ending within six hours are taken over these runs. A run's \emph{hours left} are 12 minus its run length, reported as a median over these runs. A run \emph{ends with at least two hours left} when its length is below ten hours.

\paragraph{Runs with version timelines.} \emph{Versions per run} counts the saved versions other than the starter solution, reported as a median over these runs and, in \Cref{fig:runuse}(a), as the share of runs in each range of counts.

\paragraph{Runs evaluated at every checkpoint.} These are the runs evaluated at all five checkpoints, at 1, 2, 4, 8, and 12 hours, where the 12-hour checkpoint is the final solution and each earlier checkpoint evaluates the last version saved by then. With $B_h$ the score at checkpoint $h$, clipped to $[0, 100]$ as every score is, the \emph{margin by hour} of a pair of agents is the mean of $B_h - E$ over the tasks on which the runs of both agents are evaluated at every checkpoint, with a 95\% bootstrap interval over those tasks. The \emph{share of the final score} reached by hour $h$ is the mean of $B_h$ divided by the mean of $B_{12}$ over the same tasks. The \emph{final solution is the best of the evaluated ones} when $B_{12} \ge \max(B_1, B_2, B_4, B_8)$, a comparison among the checkpoints only and not among every version that the run saved. Otherwise the \emph{shortfall} is $\max(B_1, B_2, B_4, B_8) - B_{12}$, and its share above five points is taken over all these runs. A run still working at eight hours, one whose run length exceeds eight hours, \emph{rises late} when $B_{12} > B_8$, or by more than five points when $B_{12} - B_8 > 5$, and these shares are taken over the runs still working at eight hours.

\paragraph{Development scores.} Those quoted for single runs are the task's own metric on the agent's own validation data, as recorded in its experiment log, and they are not scores on the REE scale.

\paragraph{Two special upper bounds.} One of the computed bounds of \Cref{sec:scale} comes from the development instances and lies above every score on its task. On one task, whose cost falls about 40-fold from the starter solution to a proven positive bound, $v$ in \Cref{eq:scale} is the logarithm of the cost.

\section{Per-Task Scores}
\label{app:pertask}

\begin{figure}[!ht]
\centering
\includegraphics[width=\linewidth]{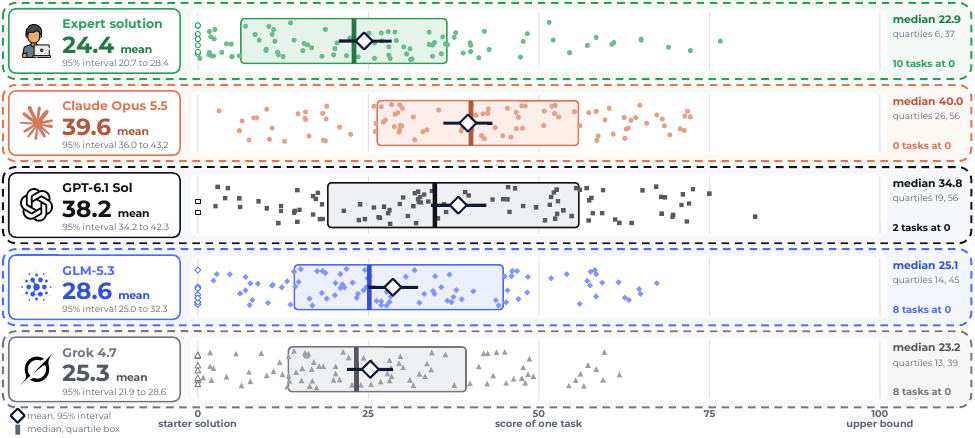}
\caption{At every quartile, the scores of Claude Opus 5.5 and GPT-6.1 Sol lie above the scores of the expert solutions. Each dot is the score of one task, from 0 (the starter solution) to 100 (the upper bound). Each box spans the first to third quartile of a row, with the median as a heavy line, and the diamond shows the mean on its 95\% bootstrap interval over tasks. The right column gives the median, the quartiles, and the number of tasks scored 0.}
\label{fig:distribution}
\end{figure}

\Cref{fig:distribution} shows the per-task scores behind the means of \Cref{tab:main}.

\end{document}